%% file: main.tex
\documentclass{article}

\usepackage[preprint]{neurips_2026}
\usepackage[utf8]{inputenc}
\usepackage[T1]{fontenc}
\usepackage[hidelinks]{hyperref}
\usepackage{url}
\usepackage{booktabs}
\usepackage{amsmath}
\usepackage{amssymb}
\usepackage{amsfonts}
\usepackage{nicefrac}
\usepackage{microtype}
\usepackage{xcolor}
\usepackage{graphicx}
\usepackage{tabularx}
\usepackage{array}
\usepackage{tikz}
\usetikzlibrary{arrows.meta,positioning,fit}

\newcommand{\method}{Q-Steer}
\newcommand{\pavsq}{PAVS-Q}
\newcommand{\E}{\mathbb{E}}
\newcommand{\Vocab}{\mathcal{V}}

\title{Q-Steer: Action-Value Guidance for Molecular Policy Optimization}

\author{
Xinyu Wang\textsuperscript{1} \quad
Jinbo Bi\textsuperscript{1} \quad
Minghu Song\textsuperscript{2}\\
\textsuperscript{1}Department of Computer Science and Engineering,
University of Connecticut\\
\textsuperscript{2}Institute of Health and Medicine,
Hefei Comprehensive National Science Center\\
\textsuperscript{1}Storrs, CT 06269, USA \quad
\textsuperscript{2}Hefei 230601, China
}

\date{}

\begin{document}
\maketitle

\begin{abstract}
Oracle-limited molecular optimization gives reward only after a complete molecule is generated, while each rollout requires many local next-token decisions. This delayed-feedback interface makes molecular policy optimization myopic: an optimizer can learn that a molecule was good without knowing which intermediate actions made it good. We introduce \method, a rollout-time action-value steering primitive for molecular language models. \method\ uses an offline-trained and frozen prefix-action value scorer, \pavsq, that estimates the downstream reward of taking a candidate next token under a partial SMILES prefix, then adds a normalized value bonus to sampling logits. The optimizer update rule and online oracle budget are unchanged; the claim is fixed-online-oracle performance, not equal total compute. On PMO23 with a fixed 10,000-call online budget, complete factorial studies across two molecular language-model backbones and four optimizers show that \method\ improves mean valid-unique score in all eight backbone--optimizer cells, with positive macro mean-score gains between +0.033 and +0.049 and 18--20 task wins per cell. Mechanism controls show that action identity matters: prefix-broadcast values are nearly neutral, while shuffled action values harm performance. These results support \method\ as a reusable rollout-time action-value wrapper that improves average molecular optimization reward across optimizer families and policy backbones without changing the online oracle budget.
\end{abstract}

\section{Introduction}

Goal-directed molecular generation is usually evaluated under a strict oracle budget: the algorithm may query a property oracle only a limited number of times, and each query requires a completed valid molecule. This makes molecular policy optimization a delayed-feedback sequence problem. A SMILES policy must choose one token at a time, but the reward that drives learning arrives only after termination. Reinforcement learning methods such as REINVENT and PPO can optimize this interface \citep{olivecrona2017molecular,blaschke2020reinvent,schulman2017ppo}, but their rollout policies remain locally myopic: at a prefix, the generator often lacks action-level information about which next token will lead to a high-scoring molecule.

The standard ways to improve this setting are to change the generator, change the optimizer, or spend the oracle budget differently. We study a more modular alternative: keep the molecular optimizer and online oracle budget fixed, but make each rollout action less myopic. If a frozen auxiliary model can estimate the future reward of each candidate next token, then the optimizer can receive local lookahead guidance while still updating exactly as before after the completed molecule is scored.

We propose \method, a rollout-time action-value steering primitive for molecular policy optimization. The guide is \pavsq, a prefix-action value scorer. It is the action-resolved counterpart of a PAVS-style prefix future-value prior: rather than only estimating whether a partial molecule is promising, it estimates which next token is promising from that prefix. Given a prefix $x_{<t}$ and a candidate token $a \in \Vocab$, \pavsq\ estimates the downstream molecular reward obtained by taking $a$ next. During sampling, \method\ shifts the optimizer's action logits by a normalized value bonus:
\begin{equation}
    \tilde{\ell}_t(a) = \ell_t(a) + \beta Q_\phi(x_{<t}, a),
    \label{eq:steer}
\end{equation}
where $\ell_t(a)$ are the original sampling logits, $Q_\phi$ denotes the centered and scaled \pavsq\ score over valid actions at the current prefix, and $\beta$ is a steering strength. The intervention is deliberately small. Within each paired comparison, the policy backbone, optimizer update rule, PMO oracle, and 10,000-call online budget remain fixed; only the rollout distribution receives action-level future-value guidance.

The design deliberately separates the claim from broader architectural or optimizer changes. \method\ is not a post-hoc reranker: it acts before each token is sampled. It is not a new PPO loss or a new REINVENT objective: completed molecules and rewards are returned to the same optimizer update. It is also not intended as a diversity mechanism. The supported claim is narrower and testable: an offline value prior can improve the average score of valid unique molecules at the same online oracle budget. This is not an equal-total-compute claim; offline \pavsq\ training is disclosed separately, and the frozen guide receives no molecules, rewards, or gradients from the evaluated online trajectory.

We test this claim with complete PMO23 factorial studies across two molecular language-model backbones and four optimizer families. Adding \method\ improves mean valid-unique score in all eight backbone--optimizer cells under the same 10,000-call online budget, with 18--20 task wins per cell. Mechanism controls close the loop: prefix-broadcast values are nearly neutral, shuffled action values harm performance, and beta sweeps show the observed reward-diversity tradeoff. The resulting story is not that \method\ universally improves every metric. Rather, it is a reusable exploitation primitive: it concentrates search using action-level future values, improving average reward while reducing uniqueness and leaving top-k discovery mixed.

\begin{figure}[t]
\centering
\includegraphics[width=\linewidth]{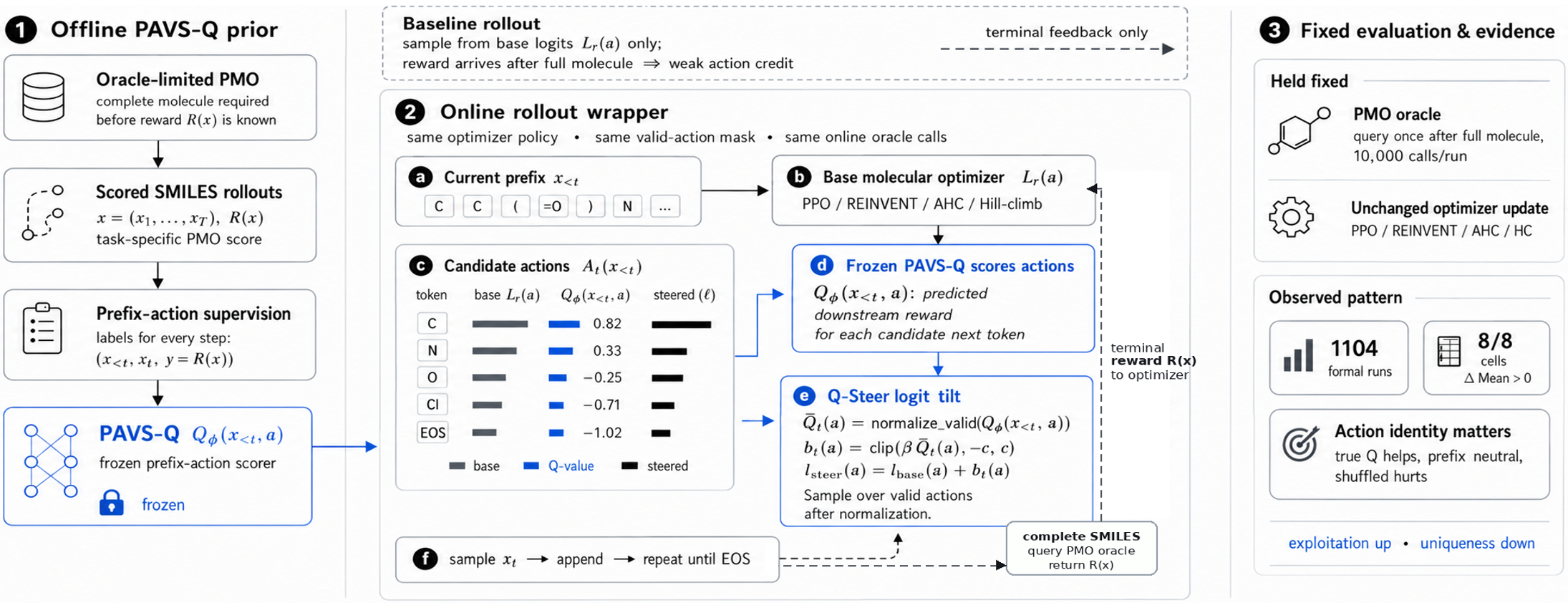}
\caption{Conceptual overview. Baseline molecular optimizers receive terminal oracle feedback after a full SMILES string is sampled, leaving prefix actions locally myopic. \method\ changes only the rollout distribution: the same optimizer proposes logits over the same valid action set, an offline-trained frozen \pavsq\ guide scores those candidate actions, and Q-Steer adds the bounded logit bonus from Equations~\ref{eq:qsteer_bonus}--\ref{eq:qsteer_logits} before the completed molecule is scored once by the PMO oracle.}
\label{fig:overview}
\end{figure}

\paragraph{Contributions.}
We make three contributions. First, we formulate \method\ as a rollout-time action-value wrapper: a frozen \pavsq\ guide scores candidate next tokens and tilts sampling logits without changing the optimizer update or online oracle budget. Second, we evaluate the wrapper in a paired PMO23 factorial study spanning four optimizer families and two backbones, showing consistent average valid-unique reward gains in all eight backbone--optimizer cells. Third, we provide mechanism and limitation evidence: action identity matters, shuffled values hurt, stronger guidance trades reward for uniqueness, and upper-tail discovery is mixed rather than universally improved.

\section{Related Work}

\paragraph{Reinforcement learning for molecular generation.}
Molecular RL methods optimize a generator against property oracles, often using SMILES language models as policies \citep{olivecrona2017molecular,blaschke2020reinvent,gottipati2020learning}. REINVENT-style approaches regularize an agent toward a prior while increasing the likelihood of high-scoring molecules, while PPO-style approaches optimize a clipped policy-gradient objective over generated trajectories \citep{schulman2017ppo}. Augmented hill-climb and related likelihood-ratio updates provide simpler exploitation-oriented alternatives that repeatedly bias generation toward high-scoring samples \citep{blaschke2020reinvent}. These methods are flexible because the oracle can encode similarity, QSAR activity, MPO objectives, or structural constraints. Their common limitation is delayed feedback: the reward is attached to a completed molecule, not to the local token actions that produced it. \method\ is designed to address this local decision problem while keeping the optimizer fixed.

\paragraph{Value guidance and credit assignment.}
Value functions are a standard solution for delayed reward in reinforcement learning, but the way they are used matters. A state value can tell whether a prefix is promising, but it does not distinguish among next-token actions from that prefix. A terminal reward model or reranker can score complete molecules, but it does not intervene during sampling. \method\ uses a prefix-action value, which is closer to a molecular analogue of action-value guidance: it estimates which next token is promising under a specific partial molecule. The prefix-broadcast and shuffled-action controls are included to test this distinction directly.

\paragraph{Decoding-time steering.}
Language-model steering methods such as PPLM, GeDi, DExperts, and FUDGE modify token probabilities at generation time using auxiliary models or discriminators \citep{dathathri2020plug,keskar2019ctrl,krause2021gedi,liu2021dexperts,yang2021fudge}. \method\ shares the idea of changing a sampling distribution without retraining the base generator, but differs in both target and signal: the guide is a task-specific molecular prefix-action value model, and the controlled variable is online PMO reward under a fixed oracle budget rather than text attributes. Unlike generic attribute steering, the guide ranks candidate molecular actions by predicted downstream PMO value and is tested by paired optimizer runs rather than by unconditional generation quality. This connection motivates treating $\beta$ as a guidance-strength parameter rather than as a new optimizer objective.

\paragraph{Oracle-budget benchmarks.}
Oracle-limited molecular optimization benchmarks emphasize fixed budgets, making it important to report not only the final best molecule but also distributional quality and search dynamics \citep{brown2019guacamol,gao2022sample}. Mean score, top-10 reward, cumulative top-10 AUC, and uniqueness measure different aspects of optimization. A method may improve average reward while reducing diversity or failing to improve early top-k discovery. Our evaluation follows this broader view rather than optimizing one headline metric.

\section{Method}

\subsection{Delayed-Reward Molecular Optimization}

Let $\pi_\theta$ be a molecular language-model policy over token sequences $x=(x_1,\ldots,x_T)$, where each token is sampled from a finite vocabulary subject to a SMILES validity mask. A black-box oracle returns a scalar reward $R(x)$ only after the full molecule has been generated and validated. Under an online oracle budget $B$, a molecular optimizer induces a sequence of rollout distributions and receives at most $B$ terminal scores. Its generic objective can be written as
\begin{equation}
    \max_\theta \; \E_{x\sim \pi_\theta}\left[R(x)\right],
    \label{eq:terminal_objective}
\end{equation}
but the policy must still choose a local action $a=x_t$ at every prefix $x_{<t}$ before observing $R(x)$. This mismatch is the credit-assignment bottleneck targeted by \method. We do not replace the optimizer that updates $\pi_\theta$; instead, we modify only the sampling distribution used to collect complete molecules.

\subsection{From PAVS to PAVS-Q}

The useful object for delayed-feedback generation is a future-value prior over partial sequences. A prefix-level value model, which we refer to as PAVS, estimates
\begin{equation}
    V_\phi(x_{<t}) \approx \E\left[R(x) \mid x_{<t}\right].
    \label{eq:pavs_v}
\end{equation}
This is informative for deciding whether a partial molecule is promising, but it does not by itself tell the generator which next token to choose. If the same scalar $V_\phi(x_{<t})$ is added to every valid action logit, the relative probabilities of actions at that prefix are unchanged.

\pavsq\ is the action-resolved version of this prior. It estimates the value of taking a specific candidate action under a prefix:
\begin{equation}
    Q_\phi(x_{<t}, a) \approx \E\left[R(x) \mid x_{<t}, x_t=a\right].
    \label{eq:pavs_q}
\end{equation}
Thus PAVS provides the future-value framing, \pavsq\ provides token-level action discrimination, and \method\ is the rollout-time wrapper that uses these action values to steer sampling. The distinction is important: $V(x_{<t})$ ranks prefixes, while $Q(x_{<t},a)$ ranks the next actions available from the same prefix.

\subsection{Training the Prefix-Action Value Model}

\pavsq\ is trained offline from scored molecular rollouts. Given a scored molecule $x=(x_1,\ldots,x_T)$ with terminal reward $R(x)$, we create supervised prefix-action examples
\begin{equation}
    \mathcal{D}_Q = \left\{\left(x_{<t}, x_t, y(x)\right): t=1,\ldots,T\right\},
    \label{eq:pavsq_dataset}
\end{equation}
where $y(x)$ is the standardized terminal reward for the PMO task. The value model is trained by mean-squared error,
\begin{equation}
    \min_\phi \; \E_{(x_{<t},a,y)\sim \mathcal{D}_Q}
    \left(Q_\phi(x_{<t},a)-y\right)^2.
    \label{eq:pavsq_loss}
\end{equation}
In our implementation, \pavsq\ uses a one-layer GRU prefix encoder, an action embedding, and an MLP head. The checkpoint is selected by held-out prefix-action validation loss and then frozen before online optimization begins. The tokenizer and action vocabulary are aligned with the online ACEGEN policy, so the action ID scored by \pavsq\ is the same action ID sampled by the optimizer. This alignment is necessary: a value model over a different tokenization would not define a valid action-level logit bonus.

For multi-component PMO objectives, \pavsq\ can estimate the scalar reward after the task-specific objective has been applied, or estimate reward components that are combined according to the PMO objective. The experiments use task-specific frozen checkpoints. This offline value-model training is an additional prior-building cost, analogous to using a pretrained molecular policy prior. It is not counted as part of the 10,000-call online PMO budget, and no generated molecules from the evaluated online trajectory are used to update \pavsq.

\subsection{Q-Steer: Rollout-Time Action-Value Steering}

At prefix $x_{<t}$, the base optimizer produces logits $\ell_t(a)$ over valid next-token actions $a\in\mathcal{A}(x_{<t})$. \method\ evaluates \pavsq\ on the same candidate action set and converts raw values into a local normalized bonus,
\begin{equation}
    \bar{Q}_t(a) = \frac{Q_\phi(x_{<t},a)-\mu_t}{\sigma_t+\epsilon},
    \qquad
    \mu_t = \frac{1}{|\mathcal{A}_t|}\sum_{a\in\mathcal{A}_t} Q_\phi(x_{<t},a),
    \label{eq:q_normalize}
\end{equation}
where $\sigma_t$ is the standard deviation over valid actions and $\mathcal{A}_t=\mathcal{A}(x_{<t})$. We then form a bounded value bonus,
\begin{equation}
    b_t(a) = \mathrm{clip}\left(\beta\bar{Q}_t(a), -c, c\right),
    \qquad a\in\mathcal{A}_t,
    \label{eq:qsteer_bonus}
\end{equation}
and add it to the base logits,
\begin{equation}
    \tilde{\ell}_t(a) = \ell_t(a) + b_t(a),
    \qquad a\in\mathcal{A}_t,
    \label{eq:qsteer_logits}
\end{equation}
with invalid actions masked before sampling. The scalar $\beta$ controls guidance strength, and $c$ bounds the final logit bonus so that rare value outliers cannot dominate the base policy. In the main experiments, $\beta=0.5$ and $c=5.0$.

The completed molecule is then scored by the PMO oracle and handed back to the underlying optimizer exactly as in the baseline. For PPO, the PPO loss and advantage objective are unchanged. For REINVENT, AHC, and hill-climbing, the likelihood update or sample-selection rule is unchanged. \method\ therefore changes the behavior policy used during rollout, but it does not introduce a new optimizer objective or additional online oracle calls.

\begin{center}
\begin{minipage}{0.94\linewidth}
\footnotesize
\hrule height 0.8pt
\vspace{0.25em}
\noindent\textbf{Algorithm 1} \quad \method\ rollout wrapper
\vspace{0.25em}
\hrule height 0.4pt
\vspace{0.30em}
\noindent\textbf{Input:} base logits $\ell_t$, frozen \pavsq\ $Q_\phi$, guidance strength $\beta$, PMO oracle budget $B$.
\vspace{0.16em}
\renewcommand{\arraystretch}{0.86}
\noindent\begin{tabularx}{0.98\linewidth}{@{}r@{\hspace{0.40em}}X@{}}
1 & \textbf{for} each online molecule until budget $B$ is exhausted \textbf{do} \\
2 & \quad Initialize prefix $x_{<1}$. \\
3 & \quad \textbf{while} generation has not terminated \textbf{do} \\
4 & \quad\quad Get base logits $\ell_t(a)$ and valid set $\mathcal{A}_t$ from the unchanged policy. \\
5 & \quad\quad Evaluate $Q_\phi(x_{<t},a)$ for actions $a\in\mathcal{A}_t$. \\
6 & \quad\quad Normalize values, clip the logit bonus, and form $\tilde{\ell}_t(a)$ by Equations~\ref{eq:qsteer_bonus}--\ref{eq:qsteer_logits}. \\
7 & \quad\quad Sample $x_t\sim\mathrm{softmax}(\tilde{\ell}_t)$ and append it to the prefix. \\
8 & \quad \textbf{end while} \\
9 & \quad Score the molecule once and pass the trajectory/reward to the original optimizer update. \\
10 & \textbf{end for}
\end{tabularx}
\vspace{0.05em}
\hrule height 0.8pt
\end{minipage}%
\end{center}

\subsection{Connection to Doob-Style Value Transforms}

The logit tilt in Equation~\ref{eq:qsteer_logits} can be viewed as a practical finite-vocabulary value transform. If one had an exact positive desirability function $h(x_{<t})$ for eventual success, a Doob $h$-transform would bias transitions by a ratio of future desirabilities, increasing the probability of trajectories that reach high-value events. In token generation, an action-resolved analogue would favor actions whose continuations have higher conditional value than other actions at the same prefix.

\method\ should not be read as a theorem-level Doob transform. \pavsq\ is learned, task-specific, and used with normalization and clipping rather than as an exact harmonic function or exact success probability. The connection is nevertheless useful because it explains the design: local transition probabilities are tilted by a frozen estimate of downstream value, while the expensive terminal oracle is still queried only after a full molecule is produced.

\subsection{Why Action-Specific Values Matter}

The method should only work if action-specific values carry information beyond prefix quality. A prefix-broadcast value $V(x_{<t})$ cannot change the relative next-token probabilities at a fixed prefix, except through numerical side effects or interactions with clipping and masks. In contrast, $Q(x_{<t},a)$ can increase the probability of one valid action while decreasing another from the same prefix. A shuffled-action control preserves the marginal distribution of value magnitudes but breaks the mapping between actions and predicted downstream rewards. These controls test whether gains come from meaningful action-value structure rather than generic logit scaling, reward-model calibration, or reduced entropy alone.

\section{Experiments}

\subsection{PMO23 Tasks and Online Budget}

We evaluate on PMO23 with a 10,000-call online oracle budget per run \citep{gao2022sample}. PMO23 includes similarity, MPO, rediscovery, scaffold, isomer, SMARTS, QED, and QSAR tasks. Each method is run with three seeds unless otherwise noted. Appendix~\ref{app:spec} gives the full task mapping, software stack, policy priors, optimizer settings, \pavsq\ training defaults, and Q-Steer rollout configuration.

\subsection{Baselines, Backbones, and Guided Settings}

The primary study is a controlled factorial comparison across four LSTM optimizer families: PPO, REINVENT, AHC, and hill-climbing. Each family is evaluated in its baseline form and with \method\ rollout steering, using the same PMO23 tasks, seeds, policy backbone, and 10,000-call online oracle budget. We denote the main guided setting as \texttt{qsteer\_b05}, corresponding to $\beta=0.5$. We additionally run the same full factorial design with a GPT2 backbone as a robustness replication. The LSTM study remains the clean controlled setting for the main causal comparison, while GPT2 tests whether the average-reward pattern survives a transformer molecular policy.

\subsection{Metrics}

We report mean score, top-10 mean score, cumulative top-10 AUC, best score, validity, uniqueness, and runtime. Unless otherwise stated, score summaries are computed over valid unique molecules produced within the 10,000-call online budget; validity is the fraction of generated molecules marked valid, and uniqueness is the fraction of valid molecules that are unique. Mean score therefore measures average valid-unique quality rather than a duplicate-weighted or invalid-as-zero reward. Top-10 and cumulative top-10 AUC measure upper-tail discovery and early discovery dynamics.

\subsection{Implementation and Run Accounting}

All primary runs use ACEGEN-aligned tokenization so that policy action IDs and \pavsq\ action values refer to the same tokens. \pavsq\ is trained offline for each PMO task and then frozen during online optimization. During rollout, values are computed on candidate actions in chunks, centered and scaled over valid actions, clipped to a bounded logit bonus, and added to logits with $\beta=0.5$ for the main setting. The optimizer-specific update rule, reward computation, and online oracle budget are unchanged.

All online comparisons count only the molecules scored during the optimizer run and use the same 10,000-call PMO oracle budget per task and seed. \pavsq\ is trained before the online optimization run and is frozen during evaluation; it does not receive gradients, rewards, or generated molecules from the online trajectory being evaluated. Thus, \method\ changes the sampling distribution but not the online oracle budget. We report \pavsq\ training as an offline prior, analogous to using a pretrained policy prior, and separate this offline cost from the online PMO budget. The policy tokenizer and \pavsq\ tokenizer are aligned by construction, and action-value scores are applied only to the current policy action IDs.

All aggregate comparisons use the paired seed intersection between a baseline optimizer and its \method\ counterpart. Tables report the number of paired seeds used for each task, and macro summaries report how many tasks have all three paired seeds. The completed LSTM factorial design contains $4 \times 2 \times 23 \times 3=552$ successful formal runs, and the matched GPT2 replication contains another 552 successful formal runs. A run is included only when the runner exits successfully and a complete \texttt{scores.csv} exists. Infrastructure failures such as MolScore server port collisions or GPU out-of-memory errors are logged and retried rather than counted as method outcomes. Appendix~\ref{app:accounting} gives the run ledger, fixed-online-budget accounting, and paired-seed aggregation rules; Appendix~\ref{app:fulltables} records the paired seed count for each task, optimizer family, and backbone.

\section{Results}

\subsection{Main Result: 8/8 Backbone--Optimizer Cells Improve Mean Reward}

Table~\ref{tab:macro} gives the PMO23 factorial summary. The table is organized around the central causal comparison: within each backbone and optimizer family, the baseline and \method\ runs use the same tasks, seeds, and 10,000-call online oracle budget, and differ only in whether rollout logits receive action-value guidance. Across 1104 successful formal PMO23 runs, all eight backbone--optimizer cells have positive mean valid-unique reward deltas. In the controlled LSTM study, \method\ improves mean score on 78/92 optimizer--task comparisons: 19/23 tasks for PPO and hill-climbing, and 20/23 tasks for REINVENT and AHC. The matched GPT2 replication adds another 77/92 task wins, with positive macro gains for every optimizer family.

\begin{table}[t]
\centering
\caption{PMO23 paired-seed macro summary over valid unique molecules. Each row compares a baseline optimizer to the same optimizer with \method\ at the same backbone, task set, seed set, and 10,000-call online oracle budget. Deltas are \method\ minus baseline. Mean, top-10, and AUC wins count tasks with positive paired-seed deltas; higher is better except for uniqueness, where negative values indicate reduced diversity.}
\label{tab:macro}
\resizebox{\linewidth}{!}{%
\begin{tabular}{llrrrrrrrr}
\toprule
Backbone & Optimizer & Tasks & 3-seed tasks & Mean wins & Top-10 wins & AUC wins & $\Delta$Mean & $\Delta$AUC & $\Delta$Unique \\
\midrule
LSTM & PPO & 23 & 23 & 19 & 7 & 12 & +0.0343 & -0.0041 & -0.2101 \\
LSTM & REINVENT & 23 & 23 & 20 & 9 & 16 & +0.0490 & +0.0365 & -0.0176 \\
LSTM & AHC & 23 & 23 & 20 & 10 & 19 & +0.0460 & +0.0364 & -0.0164 \\
LSTM & Hill-climbing & 23 & 23 & 19 & 7 & 18 & +0.0400 & +0.0424 & -0.0927 \\
GPT2 & PPO & 23 & 23 & 18 & 15 & 17 & +0.0346 & +0.0496 & -0.0057 \\
GPT2 & REINVENT & 23 & 23 & 20 & 16 & 16 & +0.0377 & +0.0357 & -0.0054 \\
GPT2 & AHC & 23 & 23 & 20 & 15 & 17 & +0.0330 & +0.0342 & -0.0021 \\
GPT2 & Hill-climbing & 23 & 23 & 19 & 12 & 16 & +0.0440 & +0.0439 & -0.0171 \\
\bottomrule
\end{tabular}%
}
\end{table}

The LSTM rows are the cleanest controlled evidence: macro mean-score gains range from +0.0343 to +0.0490, while all four optimizer families use the same policy backbone, PMO23 tasks, seeds, and online oracle budget. This matters because the effect is not confined to PPO. The same steering layer improves mean reward when attached to REINVENT, AHC, and hill-climbing, so \method\ is better understood as a sampling-time primitive that composes with multiple molecular optimizers rather than as an optimizer-specific objective modification.

The main table also separates the strength and the boundary of the result. Mean reward is the consistent gain; top-10 wins and best-score effects are mixed, and uniqueness decreases. The mechanism test in Table~\ref{tab:controls} is therefore central: it tests whether the mean-reward improvement comes from action-specific future-value guidance rather than from generic logit perturbation or entropy reduction.

\subsection{Backbone and Optimizer Robustness}

The matched GPT2 replication tests whether the pattern is tied to an LSTM molecular policy. It preserves the same direction for every optimizer family: GPT2 PPO, REINVENT, AHC, and hill-climbing all have positive macro mean-score gains, ranging from +0.0330 to +0.0440, with 18--20 mean-score task wins. GPT2 also shows clearer cumulative top-10 AUC gains than LSTM PPO, with AUC wins on 16--17 of 23 tasks across the four optimizer families. We interpret this as backbone robustness for the average-reward claim, not as evidence that every transformer architecture or scale will behave identically.

Task-bootstrap confidence intervals and exact sign tests are reported in Appendix Table~\ref{tab:macro_uncertainty}, and the complete PMO23 task-level paired deltas for all eight backbone--optimizer comparisons are in Appendix~\ref{app:fulltables}. All eight task-bootstrap intervals for $\Delta$Mean remain positive, and all exact sign tests favor \method\ at $p\leq 0.001$. This supports the average-reward claim at the PMO23 task level rather than only at the run aggregate level.

The upper-tail metrics are more nuanced. Top-10 wins are mixed for all families, and best-score wins are weaker still. Cumulative top-10 AUC improves clearly for the GPT2 rows and for the LSTM non-PPO optimizers, but is approximately neutral for LSTM PPO. Appendix Table~\ref{tab:secondary_metrics} reports the secondary macro metrics, including top-10, best-score, validity, and uniqueness deltas. These results prevent an overbroad claim: \method\ is best described as average-reward steering, not as a universally dominant best-molecule discovery algorithm.

\subsection{Reward-Diversity Tradeoff}

The consistent reward gain comes with a consistent diversity cost. Uniqueness decreases for every optimizer family and backbone. In LSTM, the macro uniqueness deltas are -0.2101 for PPO, -0.0176 for REINVENT, -0.0164 for AHC, and -0.0927 for hill-climbing; in GPT2, the corresponding deltas are smaller but still negative. This supports the mechanism-level interpretation that \method\ concentrates probability mass around actions with high predicted future value. It should therefore be treated as an exploitation knob, not a diversity-preserving exploration mechanism.

\begin{figure}[t]
\centering
\includegraphics[width=\linewidth]{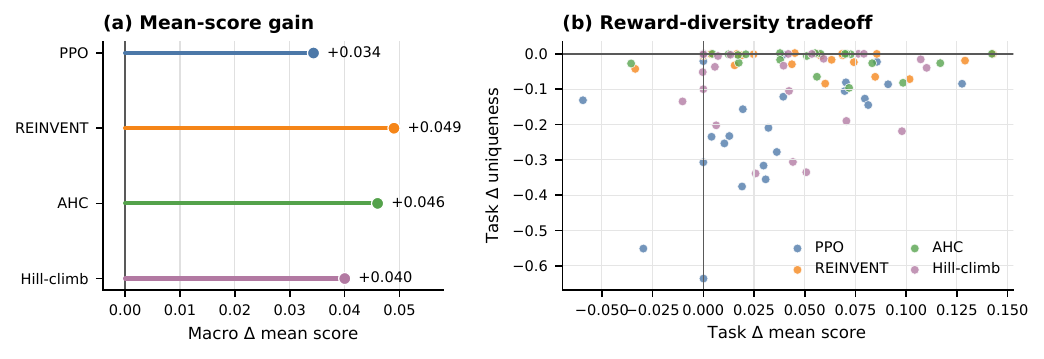}
\caption{LSTM full-factorial summary. Panel (a) shows that all four LSTM optimizer families gain macro mean reward from \method\ at the same 10,000-call online budget. Panel (b) shows PMO23 task-level paired deltas; reward gains often coincide with lower uniqueness, consistent with exploitation-oriented steering rather than diversity-preserving exploration. The matched GPT2 replication is summarized in Table~\ref{tab:macro}.}
\label{fig:factorial_diversity}
\end{figure}

Figure~\ref{fig:factorial_diversity} summarizes the LSTM macro mean-score gains across optimizer families together with the LSTM task-level reward-diversity tradeoff. The full LSTM and GPT2 task-level tables are reported in Appendix~\ref{app:fulltables}.

\subsection{Mechanism: Action-Specific Values}

The decisive ablation is not another optimizer variant; it is whether $Q(x_{<t},a)$ must be action-specific. Table~\ref{tab:controls} compares true \method\ to prefix-broadcast and shuffled-action controls in the LSTM PPO setting. True action-specific \pavsq\ improves mean score on 19/23 tasks with macro $\Delta$Mean $=+0.0343$. Prefix-broadcast values, which preserve prefix quality but remove action discrimination, are nearly neutral: 12/23 wins and $\Delta$Mean $=+0.0022$. Shuffled-action values, which preserve value magnitudes but assign them to the wrong tokens, hurt performance: 3/23 wins and $\Delta$Mean $=-0.0459$.

\begin{table}[t]
\centering
\caption{LSTM PPO mechanism controls on PMO23. Deltas are versus LSTM PPO under paired seeds. Prefix-broadcast removes action discrimination; shuffled-action breaks the mapping between values and action identities. Shuffled-action has four tasks with fewer than three paired seeds; the appendix records the partial count.}
\label{tab:controls}
\resizebox{\linewidth}{!}{%
\begin{tabular}{lrrrrrr}
\toprule
Method & Paired tasks & 3-seed tasks & Partial & Mean wins & $\Delta$Mean & $\Delta$AUC \\
\midrule
\method\ true $Q(x,a)$ & 23 & 23 & 0 & 19 & +0.0343 & -0.0041 \\
Prefix-broadcast value & 23 & 23 & 0 & 12 & +0.0022 & -0.0017 \\
Shuffled-action value & 23 & 19 & 4 & 3 & -0.0459 & -0.0633 \\
\bottomrule
\end{tabular}
}
\end{table}

This ablation is central to the story because the ordering is qualitative, not just quantitative: correct action values help, state-level broadcast values are almost inert, and wrong action assignments actively harm. The useful signal is therefore not just a score scale, logit noise, or regularizer; it is the action-specific mapping from candidate token to predicted downstream value.

We further test this interpretation with a focused beta/shuffling diagnostic on a six-task PMO subset using LSTM PPO, three seeds, and 1,500 online oracle calls per run. Table~\ref{tab:mechanism_beta} gives dose-response evidence, and Appendix Table~\ref{tab:mechanism_beta_task_appendix} reports the task-level diagnostic deltas. True action-value steering wins 5/6 tasks at each tested strength, and the macro mean gain increases from +0.0586 at $\beta=0.25$ to about +0.10 at $\beta=0.5$ and $1.0$. The cost also grows: uniqueness drops from -0.1201 to roughly -0.21 as guidance strengthens. In contrast, shuffled action values lose 0/6 tasks and reduce mean score by -0.0624. This supports the mechanism in Section~3: \method\ works by using the correct action-value assignment to concentrate rollout probability mass, not by adding arbitrary logit noise or merely reducing entropy.

\begin{table}[t]
\centering
\caption{Focused LSTM PPO mechanism diagnostic on six PMO tasks with three seeds and 1,500 online oracle calls per run. Deltas are versus PPO base. True action values improve mean score across steering strengths, while shuffled action values harm performance despite preserving value magnitudes.}
\label{tab:mechanism_beta}
\resizebox{\linewidth}{!}{%
\begin{tabular}{lrrrrrrr}
\toprule
Mode & Tasks & Mean wins & Top-10 wins & AUC wins & $\Delta$Mean & $\Delta$AUC & $\Delta$Unique \\
\midrule
True $Q$, $\beta=0.25$ & 6 & 5 & 3 & 4 & +0.0586 & +0.0630 & -0.1201 \\
True $Q$, $\beta=0.5$ & 6 & 5 & 4 & 5 & +0.1012 & +0.0998 & -0.2038 \\
True $Q$, $\beta=1.0$ & 6 & 5 & 3 & 5 & +0.1026 & +0.0927 & -0.2173 \\
Shuffled $Q$, $\beta=0.5$ & 6 & 0 & 0 & 0 & -0.0624 & -0.0639 & -0.0665 \\
\bottomrule
\end{tabular}%
}
\end{table}

\subsection{Runtime and Practical Cost}

Because \pavsq\ is an offline prior, \method\ is not compute-free: it adds model-side action scoring during rollout. Wall-clock comparisons are implementation-dependent: PPO steering is slower in our runs, while some non-PPO guided runs are faster because they were executed under different accelerated infrastructure than the original CPU-heavy baselines. Appendix Tables~\ref{tab:runtime_accounting} and~\ref{tab:compute_resources} report runtime and cluster profiles for transparency rather than as an efficiency claim. The scientific comparison is fixed-online-oracle performance; the factorial results show that extra model-side computation can improve average valid-unique reward under that budget, while runtime optimization remains necessary for deployment.

\section{Discussion}

\paragraph{Scope and next steps.}
The main result is that a frozen prefix-action value model improves average valid-unique reward across four optimizer families and two molecular language-model backbones under a fixed online oracle budget. This supports \method\ as a reusable rollout steering primitive rather than a PPO-specific modification. The GPT2 replication reduces the risk of an LSTM-only artifact, while broader transformer scaling remains future work. The cost is additional offline/model-side computation, lower uniqueness, and mixed upper-tail gains; a natural next step is to combine action-value guidance with exploration-preserving mechanisms such as novelty-aware penalties or adaptive $\beta$ schedules.

\section{Conclusion}

\method\ turns delayed molecular rewards into local action guidance. By training \pavsq\ to estimate prefix-action future values and adding bounded normalized value bonuses to rollout logits, \method\ improves average valid-unique score without changing the optimizer update or online oracle budget. Complete PMO23 factorial studies with LSTM and GPT2 backbones show robust mean-score gains across PPO, REINVENT, AHC, and hill-climbing. This supports action-value steering as a reusable primitive for oracle-limited molecular policy optimization and motivates future work on combining it with exploration-preserving mechanisms.

\section*{Broader Impacts}

This work targets oracle-limited molecular generation. It may improve sample efficiency when oracle evaluations are expensive, but the same steering mechanism could be paired with objectives that favor unsafe or undesirable molecular properties. Practical deployment should therefore use domain-specific safety filters, constrained objectives, and expert review rather than treating benchmark reward maximization as sufficient for real-world molecule design.

\section*{Limitations}

This study has several limitations. We evaluate LSTM and GPT2 backbones, but not broad transformer scaling or architecture-universal dominance. Uniqueness decreases across all optimizer families and both backbones, so \method\ should be viewed as exploitation-oriented unless paired with explicit diversity mechanisms. The method adds model-side computation because \pavsq\ must be evaluated during rollout, and \pavsq\ is a task-specific offline prior whose cost is separate from the matched online PMO budget. Top-10 and best-score improvements are mixed, so \method\ is not a universal top-k discovery enhancer. Understanding how to amortize or transfer the value model across objectives remains future work. Finally, the mechanism controls are run in LSTM PPO; extending them to every optimizer and backbone would further strengthen the mechanistic interpretation.

\newpage
{\small
\bibliographystyle{plainnat}
\bibliography{references}
}

\appendix

\section{Experimental Specification}
\label{app:spec}

This appendix provides the implementation and accounting details behind the main PMO23 factorial results. The intent is to make clear which quantities are part of the online molecular optimization budget, which quantities are offline priors, and how the paired comparisons in the main text were constructed.

\subsection{Software Stack and Benchmark Interface}

All online molecular optimization runs use ACEGEN-open as the molecular RL framework. The local ACEGEN package identifies itself as version 1.1 and depends on PyTorch/TorchRL/TensorDict, RDKit, MolScore, and PromptSMILES. The experiments use MolScore task presets to instantiate PMO and GuacaMol objectives. Each online run is launched through an ACEGEN optimizer script and writes a run record plus the ACEGEN \texttt{scores.csv} file used for aggregation.

\begin{table}[h]
\centering
\small
\caption{Software components used for the online PMO experiments.}
\label{tab:software_spec}
\resizebox{\linewidth}{!}{%
\begin{tabular}{lll}
\toprule
Component & Role in experiments & Configuration source \\
\midrule
ACEGEN-open & Molecular RL framework and optimizer scripts & \texttt{acegen-open/scripts/*/config\_denovo.yaml} \\
MolScore & PMO/GuacaMol oracle interface and task presets & \texttt{MolOpt:*} and \texttt{GuacaMol:*} task names \\
RDKit & Molecule parsing and validity/uniqueness support through MolScore & ACEGEN/MolScore dependency \\
TorchRL/TensorDict & PPO, actor, critic, rollout, and replay infrastructure & ACEGEN dependency \\
PAVS-Q package & Frozen prefix-action value scorer and Q-Steer rollout wrapper & \texttt{pavs\_value\_steering\_package} and \texttt{psv\_rl} hooks \\
\bottomrule
\end{tabular}%
}
\end{table}

\subsection{PMO23 Task Set}

The PMO23 suite spans similarity, MPO, rediscovery, scaffold, isomer, SMARTS, QED, and QSAR-style objectives. Most tasks are loaded through MolScore \texttt{MolOpt:*} presets; the three rediscovery tasks use \texttt{GuacaMol:*} presets. All formal factorial runs use a 10,000-call online oracle budget per task, seed, backbone, optimizer, and setting.

\begin{table}[h]
\centering
\scriptsize
\caption{PMO23 task names used in the factorial study.}
\label{tab:pmo23_task_list}
\resizebox{\linewidth}{!}{%
\begin{tabular}{llll}
\toprule
\texttt{albuterol\_similarity} & \texttt{amlodipine\_mpo} & \texttt{celecoxib\_rediscovery} & \texttt{deco\_hop} \\
\texttt{drd2} & \texttt{fexofenadine\_mpo} & \texttt{gsk3b} & \texttt{isomers\_c9h10n2o2pf2cl} \\
\texttt{c7h8n2o2} & \texttt{jnk3} & \texttt{median\_molecules\_1} & \texttt{median\_molecules\_2} \\
\texttt{mestranol\_similarity} & \texttt{osimertinib\_mpo} & \texttt{perindopril\_mpo} & \texttt{qed} \\
\texttt{ranolazine\_mpo} & \texttt{scaffold\_hop} & \texttt{sitagliptin\_mpo} & \texttt{thiothixene\_rediscovery} \\
\texttt{troglitazone\_rediscovery} & \texttt{valsartan\_smarts} & \texttt{zaleplon\_mpo} & \\
\bottomrule
\end{tabular}%
}
\end{table}

\subsection{Policy Backbones and Priors}

The main controlled study uses ACEGEN's default \texttt{lstm} model, which loads the ChEMBL vocabulary and \texttt{lstm\_chembl.ckpt} prior. The LSTM actor has an embedding size of 256, hidden size of 512, three recurrent layers, and no dropout in the default ACEGEN factory. The robustness replication uses ACEGEN's default \texttt{gpt2} model with the Enamine REAL vocabulary and \texttt{gpt2\_enamine\_real.ckpt} prior. The GPT2 factory uses a HuggingFace GPT2-style configuration with 24 layers, 16 attention heads, embedding size 128, context length 2048, and dropout 0.1. These priors are not retrained before the PMO runs; the optimizer updates them online according to its own algorithm.

\begin{table}[h]
\centering
\small
\caption{Policy backbone priors used in the factorial study.}
\label{tab:backbone_spec}
\resizebox{\linewidth}{!}{%
\begin{tabular}{llll}
\toprule
Backbone & ACEGEN model key & Prior checkpoint & Vocabulary/tokenizer \\
\midrule
LSTM & \texttt{lstm} & \texttt{lstm\_chembl.ckpt} & ChEMBL vocabulary, ACEGEN ChEMBL tokenizer \\
GPT2 & \texttt{gpt2} & \texttt{gpt2\_enamine\_real.ckpt} & Enamine REAL vocabulary, ACEGEN Enamine tokenizer \\
\bottomrule
\end{tabular}%
}
\end{table}

\subsection{Optimizer Hyperparameters}

Table~\ref{tab:optimizer_spec} lists the optimizer-level hyperparameters inherited from the ACEGEN denovo configurations, plus the overrides used for the PMO10k factorial runs. PPO runs disable experience replay and use one PPO epoch in the formal matrix for throughput and stability; the PPO loss, clipping objective, and advantage computation remain otherwise unchanged. For GPT2 PPO, generalized advantage estimation uses unshifted transformer sequence critics; recurrent LSTM policies keep shifted GAE under the same automatic configuration.

\begin{table}[h]
\centering
\scriptsize
\caption{Optimizer configuration used in the PMO23 factorial study. Batch parallelism was adjusted for hardware scheduling, but the online oracle budget was fixed at 10,000 scored molecules per run.}
\label{tab:optimizer_spec}
\resizebox{\linewidth}{!}{%
\begin{tabular}{lllll}
\toprule
Optimizer & Learning rate & Main optimizer parameters & Replay & PMO10k overrides \\
\midrule
PPO & $5\times10^{-4}$ & $\gamma=0.999$, $\lambda=1.0$, clip $=0.5$, critic coef. $=0.25$, entropy coef. $=0.01$, KL coef. $=0.001$, max grad norm $=0.25$ & disabled in formal PMO matrix & \texttt{ppo\_epochs=1}, \texttt{experience\_replay=false} \\
REINVENT & $1\times10^{-4}$ & $\sigma=120$ & enabled, buffer 100, replay batch 10 & default REINVENT objective unchanged \\
AHC & $1\times10^{-4}$ & top-$k=0.5$, $\sigma=60$ & enabled, buffer 100, replay batch 10 & default AHC objective unchanged \\
Hill-climbing & $1\times10^{-4}$ & top-$k=0.5$, mini-batches 2, epochs 1 & disabled & default hill-climb selection/update unchanged \\
\bottomrule
\end{tabular}%
}
\end{table}

\subsection{PAVS-Q Training and Frozen Use}

Each PMO task uses a task-specific offline \pavsq\ checkpoint. Training starts from scored SMILES data, splits molecules into train and validation sets with a 0.2 validation fraction, and decomposes every scored molecule into prefix-action examples. The supervised target is the standardized terminal task score. The default \pavsq\ architecture uses token embeddings of dimension 128, a one-layer GRU prefix encoder with hidden size 128, an action embedding projection, and an MLP head with layer normalization, GELU nonlinearities, and dropout 0.10. Training uses AdamW with learning rate $2\times10^{-3}$, weight decay $10^{-4}$, batch size 512, eight epochs, gradient clipping at 5.0, and checkpoint selection by held-out validation metric. During online PMO evaluation, the selected checkpoint is frozen and does not receive online rewards, gradients, or generated molecules from the evaluated trajectory.

\begin{table}[h]
\centering
\small
\caption{PAVS-Q value-model defaults. These are offline prior-building settings, not online PMO oracle-budget settings.}
\label{tab:pavsq_spec}
\resizebox{\linewidth}{!}{%
\begin{tabular}{ll}
\toprule
Quantity & Value \\
\midrule
Training examples & One example per molecule prefix-action pair, including EOS action \\
Target & Standardized terminal task reward \\
Train/validation split & Molecule-level split, validation fraction 0.2 \\
Prefix encoder & One-layer GRU, token embedding 128, hidden size 128 \\
Action features & Action token embedding projected to hidden size \\
MLP head & LayerNorm, GELU, dropout 0.10, scalar or component output \\
Optimization & AdamW, lr $2\times10^{-3}$, weight decay $10^{-4}$, batch size 512, 8 epochs \\
Maximum prefix length & 128 tokens \\
Online status & Frozen; no online trajectory leakage into \pavsq \\
\bottomrule
\end{tabular}%
}
\end{table}

\subsection{Q-Steer Rollout Configuration}

In the formal guided setting, the PAVS-Q hook and rollout sampling mode are enabled in the ACEGEN configuration. Table~\ref{tab:qsteer_spec} lists the rollout settings used for the main guided condition. The candidate action set is the policy's ACEGEN action vocabulary after the current valid-action mask is applied. Q-Steer changes the sampled action distribution during rollout but does not add auxiliary losses, KL anchors, reward shaping, or optimizer-specific objective changes in the formal factorial runs.

\begin{table}[h]
\centering
\small
\caption{Q-Steer rollout configuration for the formal guided PMO23 factorial setting.}
\label{tab:qsteer_spec}
\begin{tabular}{ll}
\toprule
Setting & Value \\
\midrule
Guidance strength & \texttt{rollout\_beta=0.5} \\
Control mode & \texttt{rollout\_control=action} \\
Value temperature / clip & 1.0 / 5.0 \\
Component normalization & Center and scale valid-action component scores \\
Maximum prefix length & 128 tokens \\
Score chunk / scorer batch size & 512 / 512 \\
Cache size & 50,000 prefixes \\
Auxiliary losses / reward shaping & Disabled in formal factorial runs \\
\bottomrule
\end{tabular}
\end{table}

The mechanism controls use the same machinery with different rollout controls. The prefix-broadcast control replaces all action-specific values under a prefix by the valid-action mean value, preserving prefix quality but removing action discrimination. The shuffled-action control rolls action values across the action dimension, preserving the marginal value magnitudes while breaking the action identity. The beta diagnostic uses \texttt{rollout\_beta} values 0.25, 0.5, and 1.0 on a six-task subset with 1,500 online oracle calls per run.

\subsection{Metrics and Aggregation Details}

Aggregates are computed from the ACEGEN \texttt{scores.csv} file for each successful run. The score column is selected from available task-score columns such as \texttt{score}, \texttt{single}, \texttt{gmean}, or \texttt{valid\_score}. Unless otherwise stated, mean score, top-10 score, best score, and cumulative top-10 AUC are computed over valid unique molecules. Validity is the fraction of generated rows marked valid. Uniqueness is the fraction of valid molecules marked unique. Cumulative top-10 AUC is computed by walking through available scoring steps, taking the mean of the best up to 10 valid unique scores observed so far, and averaging this curve over steps.

All comparisons are paired by task and seed. A task-level delta is the mean over paired seeds for that task. Macro deltas are then averages over PMO23 tasks, not over individual molecules. A formal run is included only if its runner record has return code 0 and a complete \texttt{scores.csv} exists. Failed infrastructure runs are retried and are not counted as method outcomes.

\subsection{Run Ledger and Online-Budget Accounting}
\label{app:accounting}

Table~\ref{tab:run_ledger} summarizes the formal run accounting used by the paper. The main factorial evidence contains 1104 successful 10,000-call PMO23 runs: 552 for the controlled LSTM study and 552 for the matched GPT2 replication. The focused beta/shuffling diagnostic is smaller by design and is used only for mechanism evidence. These counts are reported to separate completed experimental evidence from infrastructure retries and exploratory pilots.

\begin{table}[h]
\centering
\scriptsize
\caption{Formal run ledger. A successful run requires runner return code 0 and a complete ACEGEN \texttt{scores.csv}. The online oracle budget is the number of scored molecules available to the optimizer during evaluation; offline \pavsq\ training is not counted as online PMO budget.}
\label{tab:run_ledger}
\resizebox{\linewidth}{!}{%
\begin{tabular}{llllrrrl}
\toprule
Evidence block & Backbone & Optimizers / modes & Scope & Tasks & Seeds & Successful coverage & Use in paper \\
\midrule
Main PMO23 factorial & LSTM & 4 optimizers $\times$ baseline/Q-Steer & 10,000 calls & 23 & 3 & 552/552 successful runs & Primary controlled evidence \\
Backbone replication & GPT2 & 4 optimizers $\times$ baseline/Q-Steer & 10,000 calls & 23 & 3 & 552/552 successful runs & Backbone robustness \\
PMO23 mechanism controls & LSTM & PPO true-Q / prefix / shuffled controls & 10,000 calls & 23 & 3 & 23 paired tasks; shuffled has 19 full-seed and 4 partial tasks & Action-identity mechanism \\
Beta/shuffling diagnostic & LSTM & PPO base + 4 guided modes & 1,500 calls & 6 & 3 & 90/90 successful runs & Guidance-strength diagnostic \\
\bottomrule
\end{tabular}%
}
\end{table}

The paired comparisons keep the online PMO budget fixed. Q-Steer does not query the PMO oracle when scoring candidate next tokens; it evaluates only the frozen \pavsq\ guide. The completed molecule is still scored once by the same PMO oracle and then returned to the same optimizer update. Thus the scientific comparison is fixed-online-oracle performance, not equal wall-clock or equal offline-pretraining cost. This mirrors the usual use of pretrained molecular priors: offline prior construction is reported separately from the online benchmark budget.

\subsection{Secondary Metrics and Validity}

Table~\ref{tab:secondary_metrics} reports secondary macro metrics for the same paired PMO23 comparisons as Table~\ref{tab:macro}. These metrics are included to make the tradeoffs explicit. The mean-score gains are not accompanied by a uniform upper-tail improvement: top-10 and best-score effects are mixed, especially in the LSTM rows. Validity is approximately unchanged for PPO but decreases in several non-PPO settings, while uniqueness decreases in every backbone--optimizer cell. This supports the narrower interpretation used in the main text: Q-Steer is an exploitation-oriented average-reward steering method, not a diversity-preserving or universally best-molecule-improving method.

\begin{table}[h]
\centering
\scriptsize
\caption{Secondary PMO23 macro metrics under the same paired comparisons as Table~\ref{tab:macro}. Deltas are Q-Steer minus baseline, averaged over task-level paired-seed means. Wins count tasks with positive deltas.}
\label{tab:secondary_metrics}
\resizebox{\linewidth}{!}{%
\begin{tabular}{llrrrrrrrr}
\toprule
Backbone & Optimizer & Top-10 wins & $\Delta$Top-10 & Best wins & $\Delta$Best & Validity wins & $\Delta$Validity & Unique wins & $\Delta$Unique \\
\midrule
LSTM & PPO & 7 & -0.0244 & 5 & -0.0238 & 15 & -0.0006 & 0 & -0.2101 \\
LSTM & REINVENT & 9 & +0.0009 & 6 & -0.0035 & 3 & -0.0287 & 2 & -0.0176 \\
LSTM & AHC & 10 & +0.0030 & 8 & -0.0076 & 4 & -0.0252 & 4 & -0.0164 \\
LSTM & Hill-climbing & 7 & -0.0013 & 7 & -0.0167 & 3 & -0.0975 & 3 & -0.0927 \\
GPT2 & PPO & 15 & +0.0452 & 16 & +0.0413 & 6 & -0.0058 & 0 & -0.0057 \\
GPT2 & REINVENT & 16 & +0.0337 & 15 & +0.0276 & 2 & -0.0175 & 0 & -0.0054 \\
GPT2 & AHC & 15 & +0.0275 & 15 & +0.0256 & 3 & -0.0112 & 0 & -0.0021 \\
GPT2 & Hill-climbing & 12 & +0.0257 & 12 & +0.0167 & 1 & -0.1564 & 0 & -0.0171 \\
\bottomrule
\end{tabular}%
}
\end{table}

\subsection{Wall-Clock Cost Accounting}

Q-Steer adds model-side computation because candidate actions are scored by \pavsq\ during rollout. Table~\ref{tab:runtime_accounting} reports wall-clock accounting for the LSTM formal runs where runner seconds were available. These numbers should not be interpreted as a hardware-normalized efficiency benchmark: some baseline and guided non-PPO runs were executed under different scheduling and acceleration conditions. The robust conclusion is instead qualitative and conservative: Q-Steer preserves the online oracle budget but can increase model-side compute, and deployment would require engineering optimization of batched action scoring and caching.

\begin{table}[h]
\centering
\scriptsize
\caption{LSTM formal-run wall-clock accounting. Ratios are Q-Steer divided by baseline runner seconds from paired runs. Non-PPO ratios are reported for transparency but not used as scientific efficiency claims because hardware and scheduling differed across some runs.}
\label{tab:runtime_accounting}
\resizebox{\linewidth}{!}{%
\begin{tabular}{lrrrrr}
\toprule
Optimizer & Paired runs & Baseline seconds & Q-Steer seconds & Mean ratio & Median ratio \\
\midrule
PPO & 69 & 6854.1 & 11057.0 & 1.88 & 1.84 \\
REINVENT & 69 & 1760.7 & 424.7 & 0.43 & 0.12 \\
AHC & 69 & 2818.4 & 401.0 & 0.33 & 0.10 \\
Hill-climbing & 69 & 2146.8 & 521.7 & 0.48 & 0.22 \\
\bottomrule
\end{tabular}%
}
\end{table}

\subsection{Compute Resources}

The formal experiments were run on an internal Slurm cluster using the resource profiles in Table~\ref{tab:compute_resources}. Runner records store per-run wall-clock seconds, and the runtime accounting above summarizes the successful LSTM formal runs. The full research project used additional exploratory pilots and infrastructure retries; these were useful for engineering but are not counted as method outcomes unless the final runner exited successfully and produced a complete \texttt{scores.csv}. GPU jobs requested one accelerator from the cluster's \texttt{general-gpu} partition; exact accelerator model varied by node and scheduling, so runtime is reported as practical accounting rather than a hardware-normalized efficiency benchmark.

\begin{table}[h]
\centering
\scriptsize
\caption{Compute resource profiles used for the reported experiments. Time is the Slurm wall-time limit per array task; each individual PMO run also records elapsed seconds in its runner record.}
\label{tab:compute_resources}
\resizebox{\linewidth}{!}{%
\begin{tabular}{llllll}
\toprule
Experiment block & Slurm partition & Accelerator request & CPUs & Memory & Wall-time limit \\
\midrule
CPU batched PMO manifest runs & \texttt{general} & none & 20 & 240G & 12h \\
GPU batched PMO manifest runs & \texttt{general-gpu} & 1 GPU & 12 & 120G & 12h \\
CPU single-run PMO arrays / retries & \texttt{general} & none & 1 & 16G & 12h \\
Offline PAVS-Q checkpoint training & \texttt{general} & none & 1 & 32G & 12h \\
\bottomrule
\end{tabular}%
}
\end{table}

\subsection{Mechanism Diagnostic Scope}

The main mechanism controls in Table~\ref{tab:controls} use the LSTM PPO setting over PMO23. The focused beta/shuffling diagnostic in Table~\ref{tab:mechanism_beta} uses six tasks, three seeds, five modes, and 1,500 online oracle calls per run. This diagnostic is not intended to replace the full 10,000-call PMO23 factorial evidence; it is included to isolate whether action-specific value assignments, rather than generic logit perturbations, are responsible for the observed steering behavior.

\section{Full Task-Level Tables}
\label{app:fulltables}

This appendix reports the task-level paired comparisons underlying the macro summaries in the main text, along with task-bootstrap confidence intervals and exact sign tests for the main macro effects. Each row uses the paired seed intersection available for that task and comparison.

\input{appendix_tables}

\end{document}

%% file: appendix_tables.tex
% Auto-generated by experiment_scripts/generate_backbone_tables.py.
% Sources: LSTM full_factorial_qsteer/current CSVs and GPT2 pmo10k_acegen_baseline_matrix runner records.

\begin{table}[p]
\centering
\scriptsize
\caption{Full PMO23 macro summary across LSTM and GPT2 backbones. Deltas are averaged over task-level paired-seed means.}
\label{tab:full_macro_appendix}
\resizebox{\linewidth}{!}{%
\begin{tabular}{llrrrrrrr}
\toprule
Backbone & Optimizer & Tasks & Mean wins & Top-10 wins & AUC wins & $\Delta$Mean & $\Delta$AUC & $\Delta$Unique \\
\midrule
LSTM & PPO & 23 & 19 & 7 & 12 & +0.0343 & -0.0041 & -0.2101 \\
LSTM & REINVENT & 23 & 20 & 9 & 16 & +0.0490 & +0.0365 & -0.0176 \\
LSTM & AHC & 23 & 20 & 10 & 19 & +0.0460 & +0.0364 & -0.0164 \\
LSTM & Hill-climbing & 23 & 19 & 7 & 18 & +0.0400 & +0.0424 & -0.0927 \\
GPT2 & PPO & 23 & 18 & 15 & 17 & +0.0346 & +0.0496 & -0.0057 \\
GPT2 & REINVENT & 23 & 20 & 16 & 16 & +0.0377 & +0.0357 & -0.0054 \\
GPT2 & AHC & 23 & 20 & 15 & 17 & +0.0330 & +0.0342 & -0.0021 \\
GPT2 & Hill-climbing & 23 & 19 & 12 & 16 & +0.0440 & +0.0439 & -0.0171 \\
\bottomrule
\end{tabular}%
}
\end{table}

\begin{table}[p]
\centering
\scriptsize
\caption{Focused LSTM PPO beta/shuffling mechanism diagnostic on six PMO tasks with three seeds and 1,500 online oracle calls per run. Entries are task-level $\Delta$Mean versus PPO base. The scaffold task is saturated at mean score 1.0 for all modes.}
\label{tab:mechanism_beta_task_appendix}
\resizebox{\linewidth}{!}{%
\begin{tabular}{lrrrr}
\toprule
Task & True $Q$, $\beta=0.25$ & True $Q$, $\beta=0.5$ & True $Q$, $\beta=1.0$ & Shuffled $Q$, $\beta=0.5$ \\
\midrule
\texttt{albuterol\_similarity} & +0.0621 & +0.1167 & +0.1104 & -0.0928 \\
\texttt{celecoxib\_rediscovery} & +0.0445 & +0.0875 & +0.1402 & -0.0453 \\
\texttt{drd2} & +0.1673 & +0.2803 & +0.2563 & -0.1844 \\
\texttt{jnk3} & +0.0372 & +0.0830 & +0.0576 & -0.0301 \\
\texttt{qed} & +0.0404 & +0.0396 & +0.0510 & -0.0220 \\
\texttt{scaffold\_hop} & 0.0000 & 0.0000 & 0.0000 & 0.0000 \\
\bottomrule
\end{tabular}%
}
\end{table}

\begin{table}[p]
\centering
\scriptsize
\caption{Task-bootstrap uncertainty for PMO23 mean-score effects. Confidence intervals resample tasks with replacement; exact sign-test $p$ values test whether task-level mean-score deltas are symmetrically positive and negative after removing ties.}
\label{tab:macro_uncertainty}
\resizebox{\linewidth}{!}{%
\begin{tabular}{llrrrr}
\toprule
Backbone & Optimizer & Mean wins & $\Delta$Mean [95\% CI] & Sign $p$ & $\Delta$Unique [95\% CI] \\
\midrule
LSTM & PPO & 19/23 & +0.0343 [+0.0171, +0.0514] & $<0.001$ & -0.2101 [-0.2778, -0.1487] \\
LSTM & REINVENT & 20/23 & +0.0490 [+0.0317, +0.0667] & $<0.001$ & -0.0176 [-0.0284, -0.0081] \\
LSTM & AHC & 20/23 & +0.0460 [+0.0295, +0.0630] & $<0.001$ & -0.0164 [-0.0284, -0.0063] \\
LSTM & Hill-climbing & 19/23 & +0.0400 [+0.0254, +0.0552] & $<0.001$ & -0.0927 [-0.1407, -0.0489] \\
GPT2 & PPO & 18/23 & +0.0346 [+0.0187, +0.0537] & 0.001 & -0.0057 [-0.0130, -0.0014] \\
GPT2 & REINVENT & 20/23 & +0.0377 [+0.0217, +0.0565] & $<0.001$ & -0.0054 [-0.0117, -0.0012] \\
GPT2 & AHC & 20/23 & +0.0330 [+0.0174, +0.0519] & $<0.001$ & -0.0021 [-0.0039, -0.0007] \\
GPT2 & Hill-climbing & 19/23 & +0.0440 [+0.0253, +0.0659] & $<0.001$ & -0.0171 [-0.0357, -0.0045] \\
\bottomrule
\end{tabular}%
}
\end{table}

\begin{table}[p]
\centering
\scriptsize
\caption{Full PMO23 paired comparison: LSTM PPO baseline versus \method. Deltas are \method\ minus the same optimizer baseline.}
\label{tab:full_lstm_ppo}
\resizebox{\linewidth}{!}{%
\begin{tabular}{lrrrrrrr}
\toprule
Task & Seeds & Base mean & Q-Steer mean & $\Delta$Mean & $\Delta$Top-10 & $\Delta$AUC & $\Delta$Unique \\
\midrule
\texttt{albuterol\_similarity} & 3 & 0.436 & 0.468 & +0.032 & -0.001 & +0.046 & -0.209 \\
\texttt{amlodipine\_mpo} & 3 & 0.371 & 0.410 & +0.039 & -0.010 & +0.003 & -0.003 \\
\texttt{c7h8n2o2} & 3 & 0.189 & 0.130 & -0.059 & -0.208 & -0.190 & -0.131 \\
\texttt{celecoxib\_rediscovery} & 3 & 0.313 & 0.392 & +0.080 & +0.002 & +0.031 & -0.127 \\
\texttt{deco\_hop} & 3 & 1.000 & 1.000 & 0.000 & 0.000 & 0.000 & -0.307 \\
\texttt{drd2} & 3 & 0.784 & 0.815 & +0.031 & -0.000 & +0.015 & -0.355 \\
\texttt{fexofenadine\_mpo} & 3 & 0.527 & 0.563 & +0.036 & -0.004 & -0.007 & -0.278 \\
\texttt{gsk3b} & 3 & 0.446 & 0.574 & +0.128 & +0.010 & +0.031 & -0.085 \\
\texttt{isomers\_c9h10n2o2pf2cl} & 3 & 0.214 & 0.233 & +0.019 & +0.046 & +0.063 & -0.157 \\
\texttt{jnk3} & 3 & 0.164 & 0.250 & +0.086 & -0.102 & -0.053 & -0.023 \\
\texttt{median\_molecules\_1} & 3 & 0.153 & 0.192 & +0.039 & -0.035 & -0.017 & -0.121 \\
\texttt{median\_molecules\_2} & 3 & 0.165 & 0.178 & +0.013 & +0.039 & +0.038 & -0.233 \\
\texttt{mestranol\_similarity} & 3 & 0.343 & 0.425 & +0.081 & 0.000 & +0.054 & -0.145 \\
\texttt{osimertinib\_mpo} & 3 & 0.635 & 0.705 & +0.070 & +0.016 & +0.005 & -0.081 \\
\texttt{perindopril\_mpo} & 3 & 0.306 & 0.376 & +0.070 & +0.018 & +0.023 & -0.105 \\
\texttt{qed} & 3 & 0.812 & 0.831 & +0.019 & -0.000 & +0.002 & -0.376 \\
\texttt{ranolazine\_mpo} & 3 & 0.516 & 0.607 & +0.091 & -0.009 & -0.003 & -0.086 \\
\texttt{scaffold\_hop} & 3 & 1.000 & 1.000 & 0.000 & 0.000 & 0.000 & -0.021 \\
\texttt{sitagliptin\_mpo} & 3 & 0.029 & 0.033 & +0.004 & +0.008 & +0.015 & -0.235 \\
\texttt{thiothixene\_rediscovery} & 3 & 0.288 & 0.318 & +0.030 & -0.076 & -0.023 & -0.316 \\
\texttt{troglitazone\_rediscovery} & 3 & 0.245 & 0.255 & +0.010 & -0.196 & -0.092 & -0.254 \\
\texttt{valsartan\_smarts} & 3 & 0.001 & 0.001 & +0.000 & -0.001 & -0.000 & -0.636 \\
\texttt{zaleplon\_mpo} & 3 & 0.186 & 0.156 & -0.030 & -0.056 & -0.035 & -0.551 \\
\bottomrule
\end{tabular}%
}
\end{table}

\begin{table}[p]
\centering
\scriptsize
\caption{Full PMO23 paired comparison: LSTM REINVENT baseline versus \method. Deltas are \method\ minus the same optimizer baseline.}
\label{tab:full_lstm_reinvent}
\resizebox{\linewidth}{!}{%
\begin{tabular}{lrrrrrrr}
\toprule
Task & Seeds & Base mean & Q-Steer mean & $\Delta$Mean & $\Delta$Top-10 & $\Delta$AUC & $\Delta$Unique \\
\midrule
\texttt{albuterol\_similarity} & 3 & 0.266 & 0.368 & +0.102 & +0.216 & +0.253 & -0.072 \\
\texttt{amlodipine\_mpo} & 3 & 0.241 & 0.298 & +0.056 & -0.023 & -0.002 & -0.000 \\
\texttt{c7h8n2o2} & 3 & 0.107 & 0.073 & -0.034 & -0.084 & -0.007 & -0.043 \\
\texttt{celecoxib\_rediscovery} & 3 & 0.245 & 0.329 & +0.085 & -0.006 & +0.116 & -0.065 \\
\texttt{deco\_hop} & 3 & 1.000 & 1.000 & 0.000 & 0.000 & 0.000 & -0.005 \\
\texttt{drd2} & 3 & 0.344 & 0.418 & +0.074 & -0.000 & +0.025 & -0.024 \\
\texttt{fexofenadine\_mpo} & 3 & 0.384 & 0.453 & +0.069 & -0.018 & -0.008 & -0.004 \\
\texttt{gsk3b} & 3 & 0.274 & 0.403 & +0.129 & +0.047 & +0.122 & -0.019 \\
\texttt{isomers\_c9h10n2o2pf2cl} & 3 & 0.122 & 0.137 & +0.015 & -0.002 & +0.040 & -0.033 \\
\texttt{jnk3} & 3 & 0.103 & 0.148 & +0.045 & -0.066 & +0.035 & +0.002 \\
\texttt{median\_molecules\_1} & 3 & 0.107 & 0.150 & +0.044 & +0.007 & +0.039 & -0.029 \\
\texttt{median\_molecules\_2} & 3 & 0.130 & 0.146 & +0.016 & +0.039 & +0.031 & -0.000 \\
\texttt{mestranol\_similarity} & 3 & 0.227 & 0.285 & +0.058 & -0.122 & +0.067 & -0.006 \\
\texttt{osimertinib\_mpo} & 3 & 0.468 & 0.611 & +0.143 & -0.010 & +0.003 & -0.000 \\
\texttt{perindopril\_mpo} & 3 & 0.195 & 0.263 & +0.068 & -0.001 & +0.018 & +0.000 \\
\texttt{qed} & 3 & 0.667 & 0.730 & +0.063 & +0.000 & +0.003 & -0.017 \\
\texttt{ranolazine\_mpo} & 3 & 0.358 & 0.444 & +0.086 & -0.045 & -0.006 & -0.000 \\
\texttt{scaffold\_hop} & 3 & 1.000 & 1.000 & 0.000 & 0.000 & 0.000 & -0.000 \\
\texttt{sitagliptin\_mpo} & 3 & 0.014 & 0.017 & +0.004 & +0.003 & +0.014 & -0.001 \\
\texttt{thiothixene\_rediscovery} & 3 & 0.207 & 0.267 & +0.060 & +0.012 & +0.046 & -0.084 \\
\texttt{troglitazone\_rediscovery} & 3 & 0.165 & 0.190 & +0.025 & +0.088 & +0.049 & -0.001 \\
\texttt{valsartan\_smarts} & 3 & 0.000 & 0.000 & +0.000 & +0.001 & +0.001 & -0.000 \\
\texttt{zaleplon\_mpo} & 3 & 0.101 & 0.119 & +0.019 & -0.015 & -0.001 & -0.004 \\
\bottomrule
\end{tabular}%
}
\end{table}

\begin{table}[p]
\centering
\scriptsize
\caption{Full PMO23 paired comparison: LSTM AHC baseline versus \method. Deltas are \method\ minus the same optimizer baseline.}
\label{tab:full_lstm_ahc}
\resizebox{\linewidth}{!}{%
\begin{tabular}{lrrrrrrr}
\toprule
Task & Seeds & Base mean & Q-Steer mean & $\Delta$Mean & $\Delta$Top-10 & $\Delta$AUC & $\Delta$Unique \\
\midrule
\texttt{albuterol\_similarity} & 3 & 0.290 & 0.388 & +0.098 & +0.126 & +0.208 & -0.082 \\
\texttt{amlodipine\_mpo} & 3 & 0.244 & 0.302 & +0.058 & -0.007 & +0.007 & +0.000 \\
\texttt{c7h8n2o2} & 3 & 0.107 & 0.071 & -0.036 & -0.053 & +0.011 & -0.027 \\
\texttt{celecoxib\_rediscovery} & 3 & 0.274 & 0.346 & +0.072 & -0.000 & +0.142 & -0.096 \\
\texttt{deco\_hop} & 3 & 1.000 & 1.000 & 0.000 & 0.000 & 0.000 & -0.001 \\
\texttt{drd2} & 3 & 0.327 & 0.410 & +0.083 & -0.000 & +0.025 & -0.027 \\
\texttt{fexofenadine\_mpo} & 3 & 0.383 & 0.455 & +0.073 & -0.012 & -0.003 & -0.001 \\
\texttt{gsk3b} & 3 & 0.293 & 0.410 & +0.117 & +0.036 & +0.109 & -0.026 \\
\texttt{isomers\_c9h10n2o2pf2cl} & 3 & 0.119 & 0.137 & +0.017 & +0.028 & +0.033 & -0.025 \\
\texttt{jnk3} & 3 & 0.102 & 0.140 & +0.038 & -0.077 & +0.026 & +0.002 \\
\texttt{median\_molecules\_1} & 3 & 0.107 & 0.145 & +0.038 & +0.010 & +0.047 & -0.017 \\
\texttt{median\_molecules\_2} & 3 & 0.137 & 0.149 & +0.012 & +0.001 & +0.009 & -0.001 \\
\texttt{mestranol\_similarity} & 3 & 0.255 & 0.310 & +0.055 & -0.009 & +0.053 & +0.002 \\
\texttt{osimertinib\_mpo} & 3 & 0.475 & 0.618 & +0.142 & -0.008 & +0.005 & -0.000 \\
\texttt{perindopril\_mpo} & 3 & 0.196 & 0.267 & +0.071 & -0.006 & +0.012 & +0.000 \\
\texttt{qed} & 3 & 0.694 & 0.745 & +0.051 & +0.000 & +0.002 & -0.006 \\
\texttt{ranolazine\_mpo} & 3 & 0.380 & 0.450 & +0.070 & -0.039 & -0.003 & -0.000 \\
\texttt{scaffold\_hop} & 3 & 1.000 & 1.000 & 0.000 & 0.000 & 0.000 & -0.000 \\
\texttt{sitagliptin\_mpo} & 3 & 0.013 & 0.017 & +0.004 & +0.009 & +0.028 & -0.000 \\
\texttt{thiothixene\_rediscovery} & 3 & 0.222 & 0.278 & +0.056 & +0.004 & +0.049 & -0.065 \\
\texttt{troglitazone\_rediscovery} & 3 & 0.178 & 0.199 & +0.021 & +0.070 & +0.063 & -0.001 \\
\texttt{valsartan\_smarts} & 3 & 0.000 & 0.001 & +0.000 & +0.002 & +0.001 & -0.000 \\
\texttt{zaleplon\_mpo} & 3 & 0.101 & 0.118 & +0.017 & -0.006 & +0.013 & -0.003 \\
\bottomrule
\end{tabular}%
}
\end{table}

\begin{table}[p]
\centering
\scriptsize
\caption{Full PMO23 paired comparison: LSTM Hill-climbing baseline versus \method. Deltas are \method\ minus the same optimizer baseline.}
\label{tab:full_lstm_hillclimb}
\resizebox{\linewidth}{!}{%
\begin{tabular}{lrrrrrrr}
\toprule
Task & Seeds & Base mean & Q-Steer mean & $\Delta$Mean & $\Delta$Top-10 & $\Delta$AUC & $\Delta$Unique \\
\midrule
\texttt{albuterol\_similarity} & 3 & 0.293 & 0.391 & +0.098 & +0.179 & +0.234 & -0.219 \\
\texttt{amlodipine\_mpo} & 3 & 0.250 & 0.304 & +0.054 & -0.015 & +0.001 & -0.000 \\
\texttt{c7h8n2o2} & 3 & 0.060 & 0.086 & +0.026 & -0.031 & +0.060 & -0.339 \\
\texttt{celecoxib\_rediscovery} & 3 & 0.267 & 0.338 & +0.071 & -0.036 & +0.138 & -0.190 \\
\texttt{deco\_hop} & 3 & 1.000 & 1.000 & 0.000 & 0.000 & 0.000 & -0.100 \\
\texttt{drd2} & 3 & 0.186 & 0.296 & +0.110 & +0.000 & +0.026 & -0.040 \\
\texttt{fexofenadine\_mpo} & 3 & 0.424 & 0.483 & +0.059 & -0.019 & -0.008 & -0.014 \\
\texttt{gsk3b} & 3 & 0.294 & 0.401 & +0.107 & +0.074 & +0.127 & -0.016 \\
\texttt{isomers\_c9h10n2o2pf2cl} & 3 & 0.124 & 0.130 & +0.006 & +0.038 & +0.064 & -0.202 \\
\texttt{jnk3} & 3 & 0.124 & 0.166 & +0.042 & -0.020 & +0.050 & -0.000 \\
\texttt{median\_molecules\_1} & 3 & 0.120 & 0.164 & +0.044 & -0.051 & +0.034 & -0.306 \\
\texttt{median\_molecules\_2} & 3 & 0.139 & 0.139 & +0.000 & -0.022 & +0.008 & +0.000 \\
\texttt{mestranol\_similarity} & 3 & 0.224 & 0.266 & +0.042 & -0.033 & +0.128 & -0.105 \\
\texttt{osimertinib\_mpo} & 3 & 0.514 & 0.590 & +0.076 & -0.024 & -0.007 & +0.000 \\
\texttt{perindopril\_mpo} & 3 & 0.194 & 0.273 & +0.079 & +0.001 & +0.015 & +0.000 \\
\texttt{qed} & 3 & 0.710 & 0.750 & +0.040 & +0.000 & +0.002 & -0.034 \\
\texttt{ranolazine\_mpo} & 3 & 0.416 & 0.423 & +0.007 & -0.041 & -0.003 & -0.006 \\
\texttt{scaffold\_hop} & 3 & 1.000 & 1.000 & 0.000 & 0.000 & 0.000 & -0.002 \\
\texttt{sitagliptin\_mpo} & 3 & 0.017 & 0.007 & -0.010 & -0.021 & +0.021 & -0.135 \\
\texttt{thiothixene\_rediscovery} & 3 & 0.223 & 0.274 & +0.051 & +0.022 & +0.055 & -0.335 \\
\texttt{troglitazone\_rediscovery} & 3 & 0.181 & 0.195 & +0.013 & -0.025 & +0.023 & -0.003 \\
\texttt{valsartan\_smarts} & 3 & 0.001 & 0.000 & -0.000 & -0.002 & +0.000 & -0.052 \\
\texttt{zaleplon\_mpo} & 3 & 0.103 & 0.108 & +0.006 & -0.004 & +0.008 & -0.037 \\
\bottomrule
\end{tabular}%
}
\end{table}

\begin{table}[p]
\centering
\scriptsize
\caption{Full PMO23 paired comparison: GPT2 PPO baseline versus \method. Deltas are \method\ minus the same optimizer baseline.}
\label{tab:full_gpt2_ppo}
\resizebox{\linewidth}{!}{%
\begin{tabular}{lrrrrrrr}
\toprule
Task & Seeds & Base mean & Q-Steer mean & $\Delta$Mean & $\Delta$Top-10 & $\Delta$AUC & $\Delta$Unique \\
\midrule
\texttt{albuterol\_similarity} & 3 & 0.268 & 0.328 & +0.061 & +0.126 & +0.079 & -0.001 \\
\texttt{amlodipine\_mpo} & 3 & 0.280 & 0.319 & +0.039 & -0.003 & +0.003 & -0.000 \\
\texttt{c7h8n2o2} & 3 & 0.000 & 0.002 & +0.001 & +0.218 & +0.181 & -0.002 \\
\texttt{celecoxib\_rediscovery} & 3 & 0.115 & 0.197 & +0.082 & +0.161 & +0.146 & -0.006 \\
\texttt{deco\_hop} & 3 & 1.000 & 1.000 & 0.000 & 0.000 & 0.000 & -0.000 \\
\texttt{drd2} & 3 & 0.018 & 0.198 & +0.180 & +0.094 & +0.274 & -0.080 \\
\texttt{fexofenadine\_mpo} & 3 & 0.359 & 0.401 & +0.042 & -0.014 & -0.009 & -0.009 \\
\texttt{gsk3b} & 3 & 0.074 & 0.105 & +0.030 & +0.038 & +0.051 & -0.000 \\
\texttt{isomers\_c9h10n2o2pf2cl} & 3 & 0.023 & 0.096 & +0.073 & +0.133 & +0.137 & -0.013 \\
\texttt{jnk3} & 3 & 0.030 & 0.049 & +0.019 & +0.100 & +0.081 & -0.001 \\
\texttt{median\_molecules\_1} & 3 & 0.098 & 0.120 & +0.021 & +0.024 & +0.023 & -0.000 \\
\texttt{median\_molecules\_2} & 3 & 0.095 & 0.107 & +0.012 & +0.020 & +0.019 & -0.000 \\
\texttt{mestranol\_similarity} & 3 & 0.200 & 0.218 & +0.018 & +0.030 & +0.035 & -0.000 \\
\texttt{osimertinib\_mpo} & 3 & 0.426 & 0.530 & +0.104 & +0.004 & +0.011 & -0.005 \\
\texttt{perindopril\_mpo} & 3 & 0.274 & 0.263 & -0.011 & -0.003 & -0.002 & -0.000 \\
\texttt{qed} & 3 & 0.836 & 0.848 & +0.011 & -0.000 & +0.001 & -0.000 \\
\texttt{ranolazine\_mpo} & 3 & 0.026 & 0.044 & +0.018 & +0.052 & +0.044 & -0.000 \\
\texttt{scaffold\_hop} & 3 & 1.000 & 1.000 & 0.000 & 0.000 & 0.000 & -0.001 \\
\texttt{sitagliptin\_mpo} & 3 & 0.029 & 0.013 & -0.016 & -0.031 & -0.025 & -0.004 \\
\texttt{thiothixene\_rediscovery} & 3 & 0.143 & 0.196 & +0.054 & +0.067 & +0.065 & -0.001 \\
\texttt{troglitazone\_rediscovery} & 3 & 0.124 & 0.145 & +0.021 & +0.005 & +0.010 & -0.002 \\
\texttt{valsartan\_smarts} & 3 & 0.000 & 0.000 & -0.000 & -0.002 & -0.000 & -0.001 \\
\texttt{zaleplon\_mpo} & 3 & 0.086 & 0.122 & +0.036 & +0.019 & +0.021 & -0.004 \\
\bottomrule
\end{tabular}%
}
\end{table}

\begin{table}[p]
\centering
\scriptsize
\caption{Full PMO23 paired comparison: GPT2 REINVENT baseline versus \method. Deltas are \method\ minus the same optimizer baseline.}
\label{tab:full_gpt2_reinvent}
\resizebox{\linewidth}{!}{%
\begin{tabular}{lrrrrrrr}
\toprule
Task & Seeds & Base mean & Q-Steer mean & $\Delta$Mean & $\Delta$Top-10 & $\Delta$AUC & $\Delta$Unique \\
\midrule
\texttt{albuterol\_similarity} & 3 & 0.287 & 0.348 & +0.061 & +0.140 & +0.088 & -0.001 \\
\texttt{amlodipine\_mpo} & 3 & 0.265 & 0.293 & +0.028 & -0.011 & -0.003 & -0.000 \\
\texttt{c7h8n2o2} & 3 & 0.010 & 0.041 & +0.030 & +0.096 & +0.161 & -0.060 \\
\texttt{celecoxib\_rediscovery} & 3 & 0.159 & 0.229 & +0.070 & +0.058 & +0.093 & -0.024 \\
\texttt{deco\_hop} & 3 & 1.000 & 1.000 & 0.000 & 0.000 & 0.000 & -0.000 \\
\texttt{drd2} & 3 & 0.067 & 0.199 & +0.132 & +0.005 & +0.110 & -0.002 \\
\texttt{fexofenadine\_mpo} & 3 & 0.314 & 0.403 & +0.090 & -0.021 & -0.007 & -0.001 \\
\texttt{gsk3b} & 3 & 0.094 & 0.149 & +0.055 & +0.052 & +0.042 & -0.001 \\
\texttt{isomers\_c9h10n2o2pf2cl} & 3 & 0.106 & 0.139 & +0.033 & +0.010 & +0.062 & -0.009 \\
\texttt{jnk3} & 3 & 0.036 & 0.056 & +0.020 & +0.173 & +0.066 & -0.001 \\
\texttt{median\_molecules\_1} & 3 & 0.109 & 0.128 & +0.019 & +0.038 & +0.028 & -0.000 \\
\texttt{median\_molecules\_2} & 3 & 0.107 & 0.117 & +0.010 & +0.035 & +0.020 & -0.001 \\
\texttt{mestranol\_similarity} & 3 & 0.218 & 0.242 & +0.024 & +0.041 & +0.028 & -0.000 \\
\texttt{osimertinib\_mpo} & 3 & 0.377 & 0.546 & +0.169 & +0.005 & +0.012 & -0.001 \\
\texttt{perindopril\_mpo} & 3 & 0.259 & 0.267 & +0.008 & -0.018 & -0.011 & -0.000 \\
\texttt{qed} & 3 & 0.838 & 0.843 & +0.005 & -0.000 & -0.000 & -0.001 \\
\texttt{ranolazine\_mpo} & 3 & 0.043 & 0.076 & +0.033 & +0.104 & +0.070 & -0.000 \\
\texttt{scaffold\_hop} & 3 & 1.000 & 1.000 & 0.000 & 0.000 & 0.000 & -0.000 \\
\texttt{sitagliptin\_mpo} & 3 & 0.043 & 0.026 & -0.016 & -0.002 & -0.021 & -0.000 \\
\texttt{thiothixene\_rediscovery} & 3 & 0.174 & 0.224 & +0.050 & +0.058 & +0.058 & -0.018 \\
\texttt{troglitazone\_rediscovery} & 3 & 0.136 & 0.155 & +0.019 & +0.010 & +0.013 & -0.002 \\
\texttt{valsartan\_smarts} & 3 & 0.000 & 0.000 & +0.000 & +0.000 & +0.000 & -0.000 \\
\texttt{zaleplon\_mpo} & 3 & 0.120 & 0.149 & +0.029 & +0.001 & +0.012 & -0.003 \\
\bottomrule
\end{tabular}%
}
\end{table}

\begin{table}[p]
\centering
\scriptsize
\caption{Full PMO23 paired comparison: GPT2 AHC baseline versus \method. Deltas are \method\ minus the same optimizer baseline.}
\label{tab:full_gpt2_ahc}
\resizebox{\linewidth}{!}{%
\begin{tabular}{lrrrrrrr}
\toprule
Task & Seeds & Base mean & Q-Steer mean & $\Delta$Mean & $\Delta$Top-10 & $\Delta$AUC & $\Delta$Unique \\
\midrule
\texttt{albuterol\_similarity} & 3 & 0.302 & 0.355 & +0.053 & +0.096 & +0.077 & -0.001 \\
\texttt{amlodipine\_mpo} & 3 & 0.268 & 0.299 & +0.031 & -0.010 & -0.006 & -0.000 \\
\texttt{c7h8n2o2} & 3 & 0.006 & 0.017 & +0.011 & +0.069 & +0.101 & -0.016 \\
\texttt{celecoxib\_rediscovery} & 3 & 0.158 & 0.216 & +0.058 & +0.094 & +0.092 & -0.009 \\
\texttt{deco\_hop} & 3 & 1.000 & 1.000 & 0.000 & 0.000 & 0.000 & -0.000 \\
\texttt{drd2} & 3 & 0.064 & 0.188 & +0.124 & +0.006 & +0.130 & -0.001 \\
\texttt{fexofenadine\_mpo} & 3 & 0.302 & 0.385 & +0.083 & -0.020 & -0.009 & -0.001 \\
\texttt{gsk3b} & 3 & 0.090 & 0.139 & +0.049 & +0.074 & +0.051 & -0.000 \\
\texttt{isomers\_c9h10n2o2pf2cl} & 3 & 0.100 & 0.137 & +0.037 & +0.011 & +0.066 & -0.009 \\
\texttt{jnk3} & 3 & 0.034 & 0.053 & +0.019 & +0.162 & +0.112 & -0.001 \\
\texttt{median\_molecules\_1} & 3 & 0.111 & 0.124 & +0.013 & +0.021 & +0.017 & -0.000 \\
\texttt{median\_molecules\_2} & 3 & 0.108 & 0.116 & +0.008 & +0.019 & +0.013 & -0.000 \\
\texttt{mestranol\_similarity} & 3 & 0.225 & 0.246 & +0.021 & +0.025 & +0.035 & -0.000 \\
\texttt{osimertinib\_mpo} & 3 & 0.366 & 0.537 & +0.171 & +0.010 & +0.012 & -0.001 \\
\texttt{perindopril\_mpo} & 3 & 0.264 & 0.270 & +0.007 & -0.001 & +0.001 & -0.000 \\
\texttt{qed} & 3 & 0.850 & 0.853 & +0.004 & -0.000 & +0.000 & -0.000 \\
\texttt{ranolazine\_mpo} & 3 & 0.033 & 0.043 & +0.010 & +0.049 & +0.038 & 0.000 \\
\texttt{scaffold\_hop} & 3 & 1.000 & 1.000 & 0.000 & 0.000 & 0.000 & -0.000 \\
\texttt{sitagliptin\_mpo} & 3 & 0.041 & 0.028 & -0.013 & -0.018 & -0.006 & -0.000 \\
\texttt{thiothixene\_rediscovery} & 3 & 0.178 & 0.216 & +0.038 & +0.039 & +0.050 & -0.004 \\
\texttt{troglitazone\_rediscovery} & 3 & 0.140 & 0.153 & +0.013 & +0.005 & +0.007 & -0.001 \\
\texttt{valsartan\_smarts} & 3 & 0.000 & 0.000 & +0.000 & -0.001 & -0.001 & 0.000 \\
\texttt{zaleplon\_mpo} & 3 & 0.118 & 0.139 & +0.021 & +0.001 & +0.010 & -0.002 \\
\bottomrule
\end{tabular}%
}
\end{table}

\begin{table}[p]
\centering
\scriptsize
\caption{Full PMO23 paired comparison: GPT2 Hill-climbing baseline versus \method. Deltas are \method\ minus the same optimizer baseline.}
\label{tab:full_gpt2_hillclimb}
\resizebox{\linewidth}{!}{%
\begin{tabular}{lrrrrrrr}
\toprule
Task & Seeds & Base mean & Q-Steer mean & $\Delta$Mean & $\Delta$Top-10 & $\Delta$AUC & $\Delta$Unique \\
\midrule
\texttt{albuterol\_similarity} & 3 & 0.299 & 0.374 & +0.075 & +0.122 & +0.098 & -0.007 \\
\texttt{amlodipine\_mpo} & 3 & 0.267 & 0.307 & +0.040 & -0.019 & -0.014 & -0.003 \\
\texttt{c7h8n2o2} & 3 & 0.007 & 0.019 & +0.012 & +0.108 & +0.131 & -0.028 \\
\texttt{celecoxib\_rediscovery} & 3 & 0.172 & 0.254 & +0.082 & +0.053 & +0.088 & -0.037 \\
\texttt{deco\_hop} & 3 & 1.000 & 1.000 & 0.000 & 0.000 & 0.000 & -0.000 \\
\texttt{drd2} & 3 & 0.055 & 0.157 & +0.102 & +0.013 & +0.131 & -0.001 \\
\texttt{fexofenadine\_mpo} & 3 & 0.343 & 0.443 & +0.100 & -0.000 & -0.002 & -0.001 \\
\texttt{gsk3b} & 3 & 0.095 & 0.157 & +0.062 & +0.060 & +0.072 & -0.002 \\
\texttt{isomers\_c9h10n2o2pf2cl} & 3 & 0.086 & 0.110 & +0.024 & -0.013 & +0.057 & -0.018 \\
\texttt{jnk3} & 3 & 0.038 & 0.080 & +0.042 & +0.160 & +0.110 & -0.001 \\
\texttt{median\_molecules\_1} & 3 & 0.117 & 0.148 & +0.031 & +0.091 & +0.054 & -0.078 \\
\texttt{median\_molecules\_2} & 3 & 0.118 & 0.135 & +0.017 & +0.044 & +0.036 & -0.010 \\
\texttt{mestranol\_similarity} & 3 & 0.231 & 0.267 & +0.036 & +0.034 & +0.037 & -0.000 \\
\texttt{osimertinib\_mpo} & 3 & 0.408 & 0.599 & +0.191 & +0.010 & +0.010 & -0.000 \\
\texttt{perindopril\_mpo} & 3 & 0.283 & 0.296 & +0.013 & -0.004 & +0.004 & 0.000 \\
\texttt{qed} & 3 & 0.849 & 0.852 & +0.002 & -0.000 & -0.001 & -0.001 \\
\texttt{ranolazine\_mpo} & 3 & 0.114 & 0.247 & +0.134 & +0.071 & +0.195 & 0.000 \\
\texttt{scaffold\_hop} & 3 & 1.000 & 1.000 & 0.000 & 0.000 & 0.000 & -0.000 \\
\texttt{sitagliptin\_mpo} & 3 & 0.038 & 0.007 & -0.031 & -0.123 & -0.074 & -0.003 \\
\texttt{thiothixene\_rediscovery} & 3 & 0.194 & 0.238 & +0.044 & +0.033 & +0.066 & -0.184 \\
\texttt{troglitazone\_rediscovery} & 3 & 0.149 & 0.169 & +0.020 & -0.009 & +0.013 & -0.005 \\
\texttt{valsartan\_smarts} & 3 & 0.001 & 0.000 & -0.001 & -0.013 & -0.009 & -0.001 \\
\texttt{zaleplon\_mpo} & 3 & 0.116 & 0.133 & +0.017 & -0.027 & +0.004 & -0.014 \\
\bottomrule
\end{tabular}%
}
\end{table}

\begin{table}[p]
\centering
\small
\caption{PMO23 mechanism controls. Deltas are relative to LSTM PPO base under paired seeds.}
\label{tab:full_controls}
\resizebox{\linewidth}{!}{%
\begin{tabular}{lrrrrrrrr}
\toprule
Method & Tasks & 3-seed & Partial & Mean wins & Top10 wins & AUC wins & $\Delta$Mean & $\Delta$Unique \\
\midrule
True action Q & 23 & 23 & 0 & 19 & 7 & 12 & +0.034 & -0.210 \\
Prefix broadcast & 23 & 23 & 0 & 12 & 10 & 10 & +0.002 & -0.008 \\
Shuffled action Q & 23 & 19 & 4 & 3 & 1 & 2 & -0.046 & -0.139 \\
\bottomrule
\end{tabular}%
}
\end{table}